\documentclass[10pt,twocolumn,letterpaper]{article}

\usepackage{cvpr}              

\usepackage{lineno}
\usepackage{multirow}
\usepackage{bm}
\usepackage{booktabs}
\usepackage{makecell}
\usepackage[table]{xcolor}
\usepackage{graphicx}
\usepackage{amsmath,amssymb} 
\definecolor{cvprblue}{rgb}{0.21,0.49,0.74}
\usepackage[pagebackref,breaklinks,colorlinks,allcolors=cvprblue]{hyperref}

\def\paperID{****} 
\def\confName{CVPR}
\def\confYear{2026}

\title{Taming Sampling Perturbations with Variance Expansion Loss \\ for Latent Diffusion Models}

\author{Qifan Li \quad Xingyu Zhou \quad Jinhua Zhang \quad Weiyi You \quad Shuhang Gu\thanks{Corresponding author} \\
University of Electronic Science and Technology of China \\
\texttt{qifanli.lqf@gmail.com\quad shuhanggu@gmail.com}
}

\begin{document}
\maketitle
\begin{abstract}
Latent diffusion models have emerged as the dominant framework for high-fidelity and efficient image generation, owing to their ability to learn diffusion processes in compact latent spaces.
However, while previous research has focused primarily on reconstruction accuracy and semantic alignment of the latent space, we observe that another critical factor, robustness to sampling perturbations, also plays a crucial role in determining generation quality.
Through empirical and theoretical analyses, we show that the commonly used $\beta$-VAE-based tokenizers in latent diffusion models, tend to produce overly compact latent manifolds that are highly sensitive to stochastic perturbations during diffusion sampling, leading to visual degradation.
To address this issue, we propose a simple yet effective solution that constructs a latent space robust to sampling perturbations while maintaining strong reconstruction fidelity.
This is achieved by introducing a \textbf{V}ariance \textbf{E}xpansion loss that counteracts variance collapse and leverages the adversarial interplay between reconstruction and variance expansion to achieve an adaptive balance that preserves reconstruction accuracy while improving robustness to stochastic sampling.
Extensive experiments demonstrate that our approach consistently enhances generation quality across different latent diffusion architectures, confirming that robustness in latent space is a key missing ingredient for stable and faithful diffusion sampling. Our project page: {\url{https://github.com/CVL-UESTC/VE-Loss}}. 
\end{abstract}    
\section{Introduction}
\label{sec:1}

Latent diffusion models (LDMs) have become a cornerstone of modern visual generation. 
By training diffusion processes within a learned latent space rather than directly in the pixel domain, they achieve remarkable efficiency while maintaining high generative fidelity. 
Building on this principle, recent works such as \cite{rombach2022high,peebles2023scalable,ma2024sit} have achieved state-of-the-art results in high-resolution image and video synthesis, firmly establishing latent diffusion as a foundational paradigm in modern generative modeling.

Recent research has revealed that, for the tokenizer in latent diffusion models, achieving high reconstruction accuracy does not necessarily lead to better generative performance; rather, the semantic organization of the latent space also plays an crucial role \cite{yao2025reconstruction}.
%
Among these studies, VA-VAE \cite{yao2025reconstruction} align latent spaces with pretrained vision foundation models to inject semantic priors, while RAE \cite{zheng2025diffusion} go a step further by directly adopting such pretrained models as fixed tokenizers.
In contrast, works such as MAETok \cite{chen2025masked} and DC-AE 1.5 \cite{chen2025dc} leverage self-supervised learning objectives inspired by masked autoencoders \cite{he2022masked} to enhance latent representations in a data-driven manner.
%
Above approaches facilitate optimization of the diffusion process becomes easier to optimize, often exhibiting lower training losses and improved generative quality \cite{chen2025masked}. 
Although prior work has made substantial progress, what constitutes an appropriate latent space for generation remains an open question.

\begin{figure*}[ht]
    \centering
    \includegraphics[width=\textwidth]{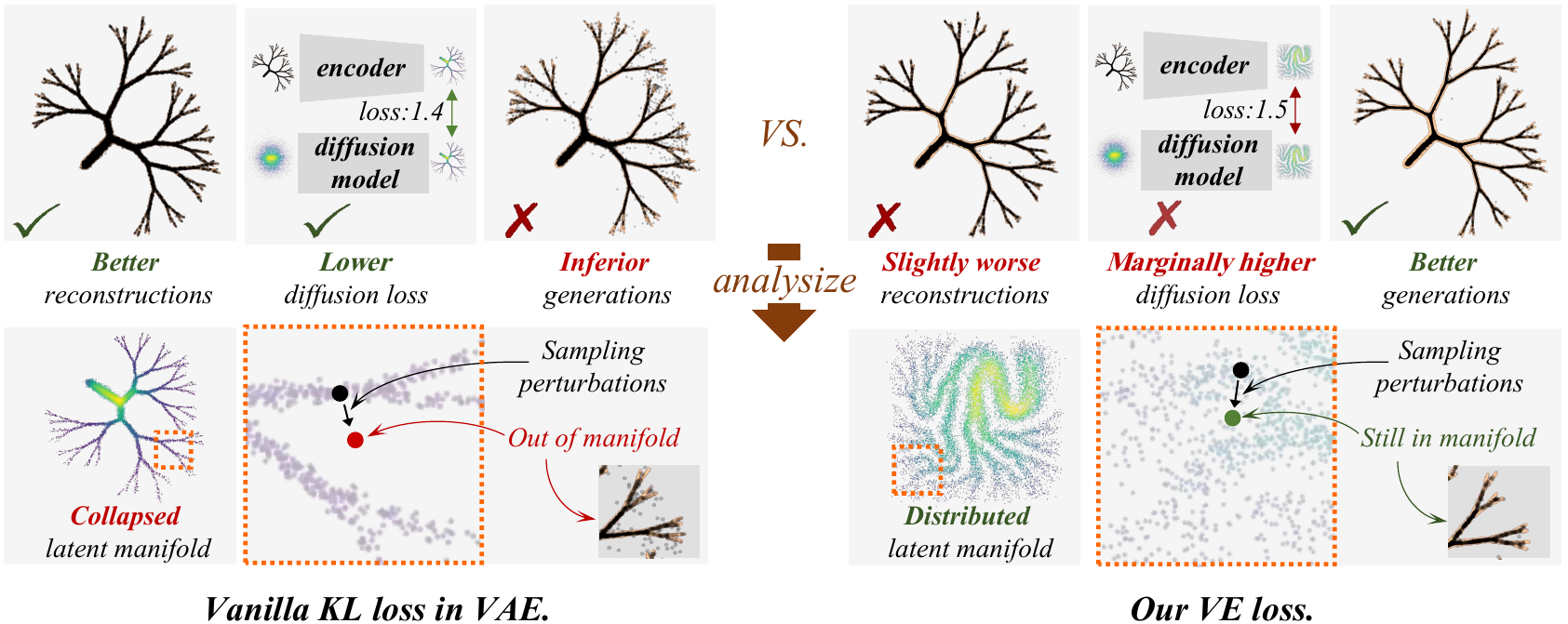}
    \caption{Toy example on a fractal-like 2D distribution following \cite{karras2024guiding}. In this toy example, we observe a seemingly counterintuitive phenomenon: a tokenizer with better reconstruction and a diffusion model trained in its latent space that achieves lower diffusion loss still produces visually inferior generations (left).
To understand this behavior, we visualize the learned latent spaces. The latent space on the left, produced by a commonly used $\beta$-VAE tokenizer, collapses into an overly compact region, which allows even small diffusion sampling perturbations to push samples outside the data manifold.
In contrast, the latent space on the right, learned with our Variance Expansion loss, remains sufficiently spread out to resist stochastic perturbations during diffusion sampling, leading to much more faithful generations.}
    \label{fig1} 
\end{figure*}

In this paper, we find that, apart from reconstruction accuracy and semantic alignment, there exists another important factor influencing generation quality.
In particular, we observe that models with near-perfect reconstruction and lower diffusion loss sometimes yield visually inferior generations compared to those with higher reconstruction errors and diffusion loss, suggesting that another property of the latent space may be at play.
As illustrated in the toy example at the top of Figure \ref{fig1},
the model on the left, which uses a vanilla tokenizer widely used in latent diffusion models, achieves higher reconstruction quality and smaller diffusion error compared to the one on the right, but its sampling results are significantly inferior.
To investigate this counterintuitive behavior, we visualize the latent representations and find that the latent manifold of the left model becomes over-compact (bottom of Figure \ref{fig1}).
Such compactness harms the robustness of the latent representation: even small stochastic perturbations during diffusion sampling can easily drive latents outside the data manifold, leading to decoding failures and degraded generation quality. 
These observations suggest that, a latent space that is robust to diffusion sampling perturbations is crucial for high-fidelity generation.

In this work, we propose a simple yet effective approach to constructing a latent space that is robust to diffusion sampling perturbations, while maintaining high reconstruction fidelity.
In particular, following the VAE formulation \cite{kingma2013auto,higgins2017beta}, we model the encoder output as a Gaussian distribution $\mathcal{N}(\mu, \sigma^2)$.
The reparameterized sampling introduces controlled randomness through $\sigma^2$, which prevents over-compact latent encoding to enhance robustness to diffusion sampling perturbations.
Building on this formulation, we further introduce a \textbf{V}ariance \textbf{E}xpansion (VE) loss, which explicitly encourages a moderate increase in $\sigma^2$ and leverages its adversarial relationship with the reconstruction loss to enable the model to learn adaptive variances $\sigma^2$.
In contrast to the commonly used KL regularization which has been shown to significantly impair reconstruction quality \cite{higgins2017beta,skorokhodov2025improving}, our method enables the model to learn adaptive $\sigma^2$ that are sufficiently large to enhance robustness against diffusion sampling perturbations while preserving reconstruction fidelity (as shown in Figure~\ref{fig2}).
Moreover, our approach preserves the learning of the latent mean $\mu$, ensuring high reconstruction quality, and remains fully compatible with recent efforts to construct discriminative latent spaces.
To summarize, our contributions are as follows:
\begin{enumerate}
\item We identify that the latent space in existing LDMs lack robustness against diffusion sampling perturbations, which often leads to degraded sampling quality.
\item We propose a simple yet effective loss function that promotes a latent space robust to diffusion sampling perturbations, while preserving strong reconstruction quality.
\item Extensive experiments conducted under various settings demonstrate that our method consistently improves generation quality, confirming its effectiveness.
\end{enumerate}

\section{Related Works}
\label{sec:3}

\subsection{Latent Diffusion Models}
Latent diffusion models train diffusion processes not directly in pixel space but in a compressed latent space learned by a pretrained autoencoder \cite{rombach2022high}.  
This design significantly reduces the computational cost of diffusion training while maintaining high perceptual quality, making it the dominant paradigm in modern visual generation models \cite{peebles2023scalable,ma2024sit,yu2024representation,wu2025representation}.
Recent research has focused on improving the semantic structure of these latent spaces.  
For example, VAVAE \cite{yao2025reconstruction} aligns the VAE latent space with pretrained semantic encoders to enhance the alignment between latent features and perceptual semantics.  
MAETok \cite{chen2025masked} and DC-AE 1.5 \cite{chen2025dc} incorporates MAE-inspired self-supervised objectives into VAE training.  
In contrast, RAE \cite{zheng2025diffusion} directly adopts a fixed vision foundation model as the tokenizer, avoiding retraining but relying entirely on the pretrained representation space.
These methods focus on improving the trainability of diffusion models by providing semantically structured latent spaces, thereby accelerating diffusion convergence and improving generative quality.  
Different from these efforts, our work investigates a complementary but largely overlooked aspect: the robustness of latent representations against diffusion sampling perturbations.
We demonstrate that, beyond trainability and semantic expressiveness, the robustness of the latent space under diffusion perturbations plays a crucial role in ensuring stable and high-fidelity generation.  

\subsection{Robustness against Sampling Perturbations}
The robustness against sampling perturbations of the latent representation strongly influences the overall stability of the generative process.
Some recent autoregressive models operating in continuous latent spaces have begun to recognize this issue.  
For instance, GIVT \cite{tschannen2024givt} mitigate this issue by strengthening the KL regularization term in the VAE tokenizer to enlarge the latent variance, while $\sigma$-VAE \cite{sun2024multimodal} adopts a fixed variance design to inject controlled stochasticity into the latent representation. 
A concurrent latent diffusion work, RAE \cite{zheng2025diffusion}, also addresses a similar problem by introducing Gaussian noise into the latent variables (analogous to a $\sigma$-VAE) to enhance robustness during generation.  
Our work shares the same insight but provides a more principled formulation: instead of heuristically adding noise, we introduce a theoretically grounded Variance Expansion Loss that adaptively balances latent robustness and reconstruction fidelity, offering a systematic solution to latent over-compact in diffusion models.

\section{Preliminary}
\label{sec:2}
In this section, we introduce the two core components of latent diffusion models, the tokenizer and the diffusion model, and describe their respective designs in detail.
\paragraph{Tokenizer.} 
A common choice of most existing latent diffusion works is a $\beta$-VAE \cite{higgins2017beta}, where an encoder $\mathcal{E}$ maps an image $\mathbf{X}_0$ to a latent distribution $\mathbf{Z}_0 \sim \mathcal{N}(\mu, \sigma^2)$. 
A sample is drawn from this distribution and passed through a decoder $\mathcal{D}$ to reconstruct the image: $\hat{\mathbf{X}} = \mathcal{D}(\mathbf{Z}_0)$.
The tokenizer is trained with a $\beta$-VAE objective, consisting of a reconstruction loss $\mathcal{L}_{\text{rec}}$ and a KL divergence term:
\begin{equation}
\mathcal{L}_{\text{VAE}} 
= \mathcal{L}_{\text{rec}} 
+ \beta \, \text{KL}(\mathcal{N}(\mu, \sigma^2) \,\|\, \mathcal{N}(0, I)).
\end{equation}
Due to the significant impact of the KL term on reconstruction quality \cite{higgins2017beta}, most latent diffusion works set $\beta$ to a very small value (e.g., $10^{-6}$) in order to prioritize reconstruction fidelity.
As a result, the latent variance $\sigma^2$ becomes negligible, making the mean $\mu$ effectively deterministic.  
Using $\mu$ or $\mathbf{Z}_0$ for diffusion training is therefore almost equivalent.

\paragraph{Diffusion model.}
In the diffusion model, the latent representation $\mu$ (or $\mathbf{Z}_0$) serves as the input.  
The forward noising process gradually corrupts these latents into $\mathbf{Z}_t$:
\begin{equation}
\mathbf{Z}_t = a(t) \mathbf{Z}_0 + b(t) \boldsymbol{\epsilon}_t, \quad \boldsymbol{\epsilon}_t \sim \mathcal{N}(\mathbf{0}, \mathbf{I}),
\end{equation}
where $a(t)$ and $b(t)$ define the noise schedule.  
The generative model, parameterized by $\boldsymbol{\theta}$, is trained to predict the added noise:
\begin{equation}
\mathcal{L}_{\text{diff}} = \mathbb{E}_{\mathbf{Z}_0, \boldsymbol{\epsilon}_t, t} \left[ \| \epsilon_{\boldsymbol{\theta}}(\mathbf{Z}_t, t) - \boldsymbol{\epsilon}_t \|^2 \right].
\end{equation}
Alternatively, flow-matching models adopt a continuous-time formulation that directly learns the vector field guiding the reverse trajectory:
\begin{equation}
\mathcal{L}_{\text{flow}} 
= \mathbb{E}_{t,\,\mathbf{Z}_t} 
\left[ \| v_{\boldsymbol{\theta}}(\mathbf{Z}_t, t) - \dot{\mathbf{Z}}_t \|^2 \right],
\end{equation}
where $v_{\boldsymbol{\theta}}$ represents the estimated velocity of the sample evolution, and $\dot{\mathbf{Z}}_t$ denotes the ground-truth time derivative of $\mathbf{Z}_t$.  
Both diffusion and flow-matching approaches share the same goal: learning the latent distribution.

\section{Methodology}
\label{sec:4}
In this section, we present a detailed analysis and describe our proposed method for improving latent robustness against sampling perturbations in diffusion models.
We begin by observing a seemingly paradoxical phenomenon: despite a tokenizer achieving high reconstruction fidelity and a diffusion model attaining low diffusion loss in its latent space, the sampling outputs are more likely to be invalid (top of Fig.~\ref{fig1}).
Further analysis reveals that this issue arises from the over-compact and sensitive latent manifolds that are easily perturbed in the diffusion sampling process (bottom of Fig.~\ref{fig1}).
Motivated by this insight, we develop a simple and effective approach to construct a robust latent space, mitigating sampling failures while preserving reconstruction quality.
\subsection{Importance of a Robust Latent Space}
\label{sec:4.1}
In this section, we use a simple toy example to demonstrate how a robust latent space against diffusion sampling perturbations plays a critical role in ensuring stable and faithful diffusion sampling.

To better understand this behavior, we construct a two-dimensional toy example following the visualization setup in \cite{karras2024guiding}.
Both the encoder and decoder operate in a two-dimensional space, allowing us to directly visualize how latent distributions evolve during training.
In this setting, we employ the most commonly used tokenizer in latent diffusion models, a $\beta$-VAE with a very small KL regularization weight, as discussed above.
We first train the tokenizer and diffusion model separately until convergence, and then analyze their interaction during sampling.
Interestingly, we observe that the latent manifold tends to become overly compact, forming a thin, needle-like manifold as shown in the bottom-left of Figure \ref{fig1}.
We argue that this phenomenon arises because, when the KL term is too weak, the encoder naturally tends to minimize latent uncertainty in order to achieve precise reconstructions through the reconstruction loss. As a result, the latent distribution collapses into near-deterministic values.
\begin{figure*}[ht]
    \centering
     \includegraphics[width=\textwidth]{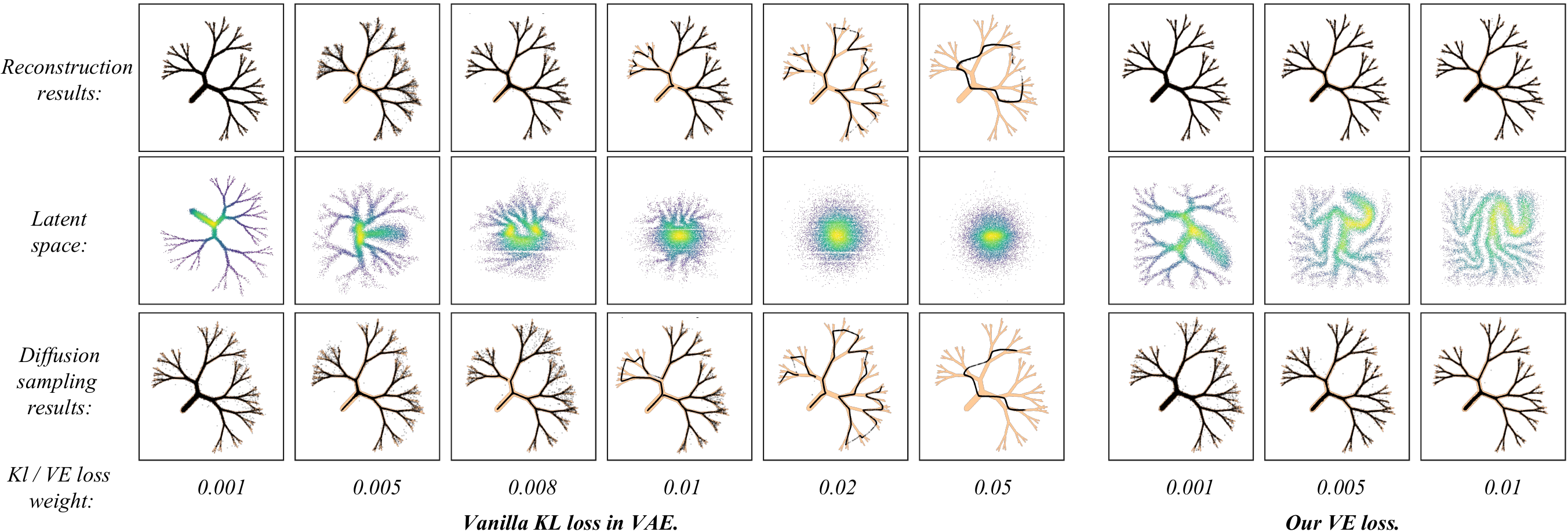}
    \caption{Toy example on a fractal-like 2D distribution following \cite{karras2024guiding}. KL regularization severely degrades reconstruction quality (left), whereas our method maintains high reconstruction quality (right).}
    \label{fig2} 
\end{figure*}
We also present some theoretical analysis of this behavior below:
\paragraph{\textit{Analyze 1: Reconstruction-induced variance collapse.}}
Let the encoder output follow a Gaussian distribution:
\begin{equation}
z \sim \mathcal{N}(\mu, \sigma^2),
\end{equation}
and let $\mathcal{D}(z)$ denote the decoder output.
For sufficiently small $\sigma$, we approximate the decoder locally via a first-order Taylor expansion around $\mu$:
\begin{equation}
\mathcal{D}(\mu + \sigma\epsilon) \approx \mathcal{D}(\mu) + J(\mu)\,\sigma\epsilon,
\end{equation}
where $J(\mu)=\left.\frac{\partial \mathcal{D}}{\partial z}\right|_{z=\mu}$ is the decoder Jacobian, and $\epsilon\sim\mathcal{N}(0,I)$.
This linearization isolates the dominant term describing how the reconstruction loss depends on the latent variance.
Although $\mathcal{D}$ is generally nonlinear, higher-order Taylor terms are at least $O(\sigma^2)$ in the expansion and therefore contribute $O(\sigma^4)$ or higher to the expected squared error (with no $O(\sigma^3)$ term appearing after expectation due to the symmetry of $\epsilon \sim \mathcal{N}(0,I)$, which causes all odd-order terms to vanish), and thus do not alter the monotonic trend that drives variance collapse.
The expected reconstruction loss (squared error) under the reparameterized sampling becomes:
\begin{equation}
\begin{aligned}
\mathcal{L}_{\mathrm{rec}}(\mu,\sigma)
&= \mathbb{E}_{\epsilon}\big[\|\mathbf{X}_0 - \mathcal{D}(\mu+\sigma\epsilon)\|^2\big] \\
&\approx \underbrace{\|\mathbf{X}_0 - \mathcal{D}(\mu)\|^2}_{\text{deterministic term}}
\;+\; \sigma^2\,\mathbb{E}_{\epsilon}\big[\|J(\mu)\epsilon\|^2\big] \\
&= \|\mathbf{X}_0 - \mathcal{D}(\mu)\|^2 + \sigma^2\,\mathrm{Tr}\big(J(\mu)J(\mu)^\top\big),
\end{aligned}
\end{equation}
where we define $T(\mu):=\mathrm{Tr}\big(J(\mu)J(\mu)^\top\big)$ as a local sensitivity measure of the decoder.
Hence, the reconstruction loss can be simplified as:
\begin{equation}
\mathcal{L}(\mu,\sigma) \approx 
\|\mathbf{X}_0 - \mathcal{D}(\mu)\|^2 + \sigma^2 T(\mu).
\label{eq:recon_taylor}
\end{equation}
This result shows that the reconstruction loss grows approximately linearly with $\sigma^2$, with slope $T(\mu)$.  
Consequently, minimizing $\mathcal{L}_{\mathrm{rec}}$ alone induces a strong gradient toward smaller $\sigma$ (since $\partial \mathcal{L}_{\mathrm{rec}}/\partial \sigma \approx 2\sigma T(\mu)$), explaining the empirical tendency of variance collapse.

This distribution collapse makes the latent space highly vulnerable during diffusion sampling, where stochastic perturbations frequently occur in generative process.
As shown in \ref{fig1}, latents with overly narrow manifolds exhibit sharp boundaries, beyond which the decoder cannot reconstruct meaningful content.
Even slight deviations from the latent manifolds can easily lead to severe degradation or complete generation failure after decoding.
This analysis highlights a fundamental issue in existing latent diffusion pipelines: the learned latent spaces are not robust to sampling perturbations.
Ensuring robustness in the latent space is therefore crucial for achieving reliable diffusion-based generation.

\subsection{Variance Expansion Loss}
\label{sec:4.2}
In this section, we aim to build a robust latent space against sampling perturbations while maintaining strong reconstruction performance.

When considering a standard VAE, the latent prior plays a central role in shaping the latent space: it explicitly constrains where encoded representations should lie and provides a statistical reference that regularizes the encoder.
However, in a latent diffusion model, this relationship changes fundamentally. Rather than generating data directly from a fixed prior, the model learns a denoising trajectory that progressively transforms Gaussian noise into meaningful latent representations.
Through this process, the diffusion model itself learns the structure of the latent distribution as part of its generative dynamics.
As a result, the KL regularization term is not a necessary component for latent diffusion models, since the encoder no longer needs to align its outputs with a predefined Gaussian prior. The diffusion process inherently discovers and enforces the prior distribution that is most suitable for generation.
In fact, the KL term can even degrade overall performance by severely impairing reconstruction quality, as discussed in \cite{higgins2017beta,skorokhodov2025improving} and illustrated in our toy example (Fig.~\ref{fig2}).
We also provide some theoretical analysis in Appendix.

Building upon these analyses, we discard the conventional KL term and instead take a more principled, goal-driven approach to loss design.
Our objective is to construct a loss function that enhances the robustness of the latent space against sampling perturbations.
As discussed in Section~\ref{sec:4.1}, this can be achieved by preventing the latent variance $\sigma^2$ from collapsing under the influence of the reconstruction loss.
Motivated by this insight, we propose a simple yet effective \textbf{V}ariance \textbf{E}xpansion loss (\textbf{VE} loss), which counteracts the collapsing tendency of the reconstruction objective and maintains a healthy latent variance, thereby improving robustness to sampling perturbations.
Specifically, we explicitly encourages moderate variance in the latent distribution:
\begin{equation}
\mathcal{L}_{\text{var}}(\sigma) = \frac{1}{\sigma^2 + \delta},
\label{eq:log_var}
\end{equation}
where a tiny $\delta>0$ ensures numerical stability. 
However, this tends to cause the overall magnitude to increase, so we introduce an empirical regularization term:
\begin{equation}
\mathcal{L}_{\text{reg}} = e^{|z| - \tau }.
\end{equation}
where $\tau$ serves as a threshold parameter determining the activation range of the exponential penalty.
The overall training objective is formulated as:
\begin{equation}
\mathcal{L} = \mathcal{L}_{\text{rec}} + \lambda_1 \,\mathcal{L}_{\text{var}} + \lambda_2\,\mathcal{L}_{\text{reg}},
\end{equation}
where $\lambda_1$ controls the trade-off between robustness and reconstruction accuracy, $\lambda_2$ controls the regularization strength and $\mathcal{L}_{\text{rec}}$ is a normal reconstruction loss following previous works \cite{esser2021taming,rombach2022high}.
Compared with the KL regularization in standard VAEs, which enforces a fixed, globally uniform Gaussian constraint regardless of the local data geometry, our variance expansion loss provides an \emph{adaptive} mechanism.  
It automatically adjusts the allowable latent variance according to the decoder’s sensitivity, enabling the model to maintain stability in regions of high curvature while allowing greater flexibility in smoother regions.
We provide some theoretical analysis of this behavior below:
\begin{table*}[h]
\centering
\caption{Comprehensive comparisons show that the VE loss consistently improves generative performance across architectures while maintaining competitive reconstruction quality. $\downarrow$ and $\uparrow$ indicate whether lower
or higher values are better, respectively. (\textbf{LightningDiT-B}$^+$ denotes the experiment with training extended to 160 epochs.)}
\renewcommand{\arraystretch}{1.1}
\resizebox{0.9\textwidth}{!}{
\begin{tabular}{lcccccc|ccc}
\toprule
\multirow{2}{*}{\textbf{Tokenizer}} & \multirow{2}{*}{\textbf{Epochs.}} & \multirow{2}{*}{\textbf{Spec.}} 
& \multicolumn{4}{c|}{\textbf{Reconstruction Performance}} 
& \multicolumn{3}{c}{\textbf{Generation Performance (FID-10K)$\downarrow$}} \\
&& & \textbf{rFID}$\downarrow$ & \textbf{PSNR$\uparrow$} & \textbf{LPIPS$\downarrow$} & \textbf{SSIM$\uparrow$} & \textbf{DiT-B} & \textbf{LightningDiT-B} & \textbf{LightningDiT-B}$^+$ \\
\midrule
LDM &10 & \multirow{2}{*}{f16d16} &\textbf{0.55}  &\textbf{26.05}  &\textbf{0.141} &\textbf{0.710}  &31.93  &22.25  & - \\
LDM+VE loss &10&&0.60  &25.23  &0.147 &0.691     &\textbf{29.03}  &\textbf{19.70}  & -  \\
\midrule
VAVAE &16& \multirow{3}{*}{f16d32} &\textbf{0.35 } &\textbf{27.43 } &\textbf{0.104}  &\textbf{0.77}  &22.27  &19.85  &-  \\
VAVAE &\textcolor{gray}{50}&  &\textcolor{gray}{0.28}  &\textcolor{gray}{27.96}  &\textcolor{gray}{0.096}  &\textcolor{gray}{0.79}  & \textcolor{gray}{-} & \textcolor{gray}{-} &\textcolor{gray}{15.82}  \\
VAVAE+VE loss  &16& &0.45  &26.54  &0.118  &0.74  &\textbf{19.42}  &\textbf{15.50}  &\textbf{12.89}  \\
\bottomrule
\end{tabular}}
\label{tab:1}
\vspace{-1ex}
\end{table*}
\paragraph{\textit{Analyze 2: Gradient Balancing and Design of the Variance Expansion Term.}}  
The reconstruction term in latent diffusion tokenizers induces a gradient on the latent variance $\sigma$ that drives it toward zero.
Specifically, by differentiating Eq.~\ref{eq:recon_taylor}, we obtain
\begin{equation}
\frac{\partial \mathcal{L}{\mathrm{rec}}}{\partial\sigma} \approx 2\sigma,T(\mu),
\end{equation}
where $T(\mu)$ measures the local sensitivity of the decoder around $\mu$.
This gradient naturally pushes $\sigma$ toward zero, leading to a collapse of the latent variance and thereby reducing the robustness of the latent space to stochastic perturbations during diffusion sampling.
To prevent such variance collapse, we introduce a variance expansion term $\mathcal{L}_{\text{var}}(\sigma)$ (see equation\ref{eq:log_var}) that applies an opposing gradient.
Let $g(\sigma) := \partial_\sigma \mathcal{L}_{\text{var}}$ denote its gradient; then, the equilibrium condition for $\sigma$ can be expressed as
\begin{equation}
2\sigma T(\mu) + g(\sigma) = 0,
\end{equation}
where $g(\sigma)$ should exert a strong counteracting force when $\sigma$ is small to resist collapse, while gradually diminishing for larger $\sigma$ to prevent over-dispersion.

We consider three natural candidate forms for $\mathcal{L}_{\text{var}}$: \textbf{(i)} Negative variance: $\mathcal{L}_{\text{var}}^{(1)}=-\alpha\sigma^2$ with $g(\sigma)=-2\alpha\sigma$. The equilibrium condition reduces to $\alpha=T(\mu)$, which does not determine a local $\sigma$ and fails to provide a self-stabilizing solution across varying $T(\mu)$. Moreover, the gradient vanishes as $\sigma\to0$, offering no protection against collapse. \textbf{(ii)} Log-entropy: $\log\sigma^2$ with gradient
$g(\sigma)\propto 1/\sigma$. This yields a non-trivial equilibrium $\sigma^2=\beta/T(\mu)$, providing a locally adaptive variance that inversely scales with decoder sensitivity. While theoretically valid, the gradient magnitude in small-$\sigma$ regions may be insufficient to fully prevent collapse. \textbf{(iii)} Inverse variance (our choice): $\mathcal{L}_{\text{var}}^{\mathrm{inv}} = \lambda/(\sigma^2+\delta)$ with $g(\sigma)=-2\lambda/\sigma^3$. 
The equilibrium satisfies
\begin{equation}
    \sigma^4 = \frac{\lambda}{T(\mu)} \quad\Longrightarrow\quad \sigma = \left(\frac{\lambda}{T(\mu)}\right)^{1/4}.
\end{equation}
This design provides a strong, locally adaptive restoring force that increases rapidly as $\sigma\to0$, reliably counteracting collapse while naturally decaying for larger $\sigma$. It thus achieves the desired trade-off between robustness and reconstruction fidelity.
Combining reconstruction and variance expansion terms, the total objective for a single latent location is:
\begin{equation}
\mathcal{L}(\mu,\sigma) \approx 
\underbrace{\|\mathbf{X}_0 - \mathcal{D}(\mu)\|^2 + \sigma^2 T(\mu)}_{\mathcal{L}_{\text{rec}}}
\;+\; \lambda \underbrace{\frac{1}{\sigma^2 + \delta}}_{\mathcal{L}_{\text{var}}}, 
\end{equation}
whose minimization yields a locally adaptive latent variance that matches the geometry of the decoder, enhancing robustness to diffusion sampling perturbations while preserving reconstruction fidelity.

Overall, through the natural adversarial relationship between the reconstruction loss, which favors smaller variance, and the \textbf{VE} loss, which encourages larger variance, the model learns adaptive variances across the latent manifold that are large enough to absorb diffusion perturbations while maintaining reconstruction fidelity. The effect can be observed in our toy example shown in Figure~\ref{fig1} and ~\ref{fig2}. The detailed setting of toy example can be found in Appendix.

\subsection{Discussion}
Previous generative modeling approaches have also employed certain tricks that, to some extent, improve the robustness of the latent space.
For instance, methods that strengthen the KL regularization in VAE-based tokenizers, which can reduce the sensitivity of the latent space to sampling perturbations \cite{tschannen2024givt}.
Similarly, other approaches inject fixed noise into the latent representation by defining a constant variance and letting the encoder output only the mean \cite{sun2024multimodal,zheng2025diffusion}.
These techniques often bring moderate improvements in and sampling robustness, even though they were originally designed for other purposes such as feature disentanglement regularization.
However, these strategies remain largely heuristic.
A stronger KL term enforces alignment with a Gaussian prior but unavoidably suppresses the expressive capacity of the latent representation, leading to blurred reconstructions, as discussed in Section~\ref{sec:4.2}.
Conversely, fixing a large global variance sidesteps this issue but relies on manually tuned, non-adaptive hyperparameters, ignoring the fact that different latent regions exhibit different sensitivities to diffusion noise.

In contrast, our approach aims to address the problem from a more principled perspective.
Rather than relying on global or manually tuned tricks, we explicitly model and optimize latent robustness in an adaptive manner, allowing the variance to self-adjust according to local decoder sensitivity.
This creates a latent space that balances reconstruction fidelity with robustness to diffusion perturbations, forming the foundation of our proposed \textbf{VE} Loss.

\section{Experiments}
\label{sec:5}
\subsection{ Implementation Details}
\paragraph{Tokenizers.}
All experiments are conducted on the \textit{ImageNet} dataset \cite{deng2009imagenet} at a resolution of $256\times256$. 
The tokenizer follows the architecture and training strategy from \cite{esser2021taming,rombach2022high,yao2025reconstruction}.
We replace the commonly used KL divergence term \cite{esser2021taming,rombach2022high,yao2025reconstruction} with our \textbf{VE} loss to encourage latent robustness against diffusion sampling perturbations. 
All tokenizers in our experiments downsample the input by a factor of 16, following the setting in~\cite{yao2025reconstruction}.
Empirically, we set the variance expansion loss weight $\lambda_1$, the regularization loss weight $\lambda_2$, and the threshold-like parameter $\tau$ to $0.1$, $1\times10^{-6}$, and $1$, respectively.
Training is performed on eight NVIDIA RTX 4090 GPUs with a global batch size of $64$, determined by the maximum available GPU memory.
All models are optimized using the AdamW optimizer \cite{kingma2014adam} with $\beta_1=0.5$, $\beta_2=0.9$ and a learning rate of $2.5\times10^{-5}$, linearly scaled from \cite{yao2025reconstruction}.
We train our tokenizer models for 16 epochs on our ablation studies. For the state-of-art one, due to limited computational resources, we fine-tune VA-VAE \cite{yao2025reconstruction} for 5 epochs. 
\paragraph{Generative Models.}
For the generative model, we adopt the flow matching objective with linear interpolation $\bm{X}_t = (1-t) \bm{X} + t\bm{\epsilon}$,
where $\bm{X} \sim p(\bm{X})$ and $\bm{\epsilon} \sim \mathcal{N}(0, \mathbf{I})$, and train the model to predict the velocity $v(\bm{X}_t, t)$ following standard practice.
We use both the vanilla setups from DiT~\cite{peebles2023scalable} and SiT~\cite{ma2024sit} (hereafter collectively referred to as DiT), as well as LightningDiT~\citep{yao2025reconstruction}, a variant of DiT, as our model backbones in ablation studies to validate the generality of our method.
For efficiency, all our ablation studies are conducted on the base model (130M).
In our ablation studies, all training is performed on eight NVIDIA RTX 4090 GPUs with a global batch size of $1024$ following the setting \cite{yao2025reconstruction}.
All models are optimized using the AdamW optimizer \cite{kingma2014adam} with $\beta_1=0.95$, $\beta_2=0.999$ and a learning rate of $2\times10^{-4}$ following \cite{yao2025reconstruction}.
For the state-of-the-art configuration, we use LightningDiT-XL (675M) \cite{yao2025reconstruction} as our generative model. 
It is worth noting that diffusion models trained in the latent space aligned with DINOv2 \cite{oquab2023dinov2} representations \cite{yao2025reconstruction,zheng2025diffusion}, despite their strong performance, are widely recognized for exhibiting training instability when optimized over long schedules.
We encountered this issue when training with the commonly used AdamW \cite{kingma2014adam} optimizer over long durations. 
In contrast, we found that the Muon optimizer \cite{jordan2024muon} effectively mitigates this problem. 
Therefore, we adopt Muon optimizer for long-term training.
We provide a analysis of this part in Appendix.
Training is performed on four NVIDIA RTX Pro 6000 GPUs with a global batch size of $768$, determined by the maximum available GPU memory.
The model is optimized with a learning rate of $1.8\times10^{-4}$, log-scaled from~\cite{yao2025reconstruction,zheng2025diffusion}, using $\beta_1=0.95$, $\beta_2=0.999$ and an EMA update rate of 0.9999.
For all experiments, we adopt a patch size of 1 for all models on on \textit{ImageNet} at $256\times256$ following \cite{yao2025reconstruction} which results in a sequence length of 256, matching the token length used by DiTs \citep{peebles2023scalable,ma2024sit} and thus maintaining the same computational cost.
\paragraph{Evaluations.} 
For tokenizers, we report the reconstruction Fréchet Inception Distance (rFID) \cite{heusel2017gans}, PSNR , LPIPS \cite{zhang2018unreasonable} and SSIM \cite{wang2004image} to assess the reconstruction quality.
For generative models, we using the Fréchet Inception Distance (FID)~\cite{heusel2017gans}, computed on 10K samples for ablation studies (denoted as FID-10K) and 50K samples for state-of-the-art models, generated with 250 sampling steps using the Euler sampler. 
In addition, we evaluate the generative quality using the Inception Score (IS)~\cite{salimans2016improved}, Precision, and Recall.
%
\begin{table}[t]
\centering
\caption{System-level comparison on \textit{ImageNet}
256×256 without classifier-free guidance (CFG). $\downarrow$ and $\uparrow$ indicate whether lower or higher values are better, respectively.}
\resizebox{0.9\columnwidth}{!}{
\begin{tabular}{l|c|cc|c}
\toprule
\multirow{2}{*}{\textbf{Method}}
&\multirow{2}{*}{$\bm{\sigma^2}$} 
& \multicolumn{2}{c}{\textbf{Reconstruction}} & \multicolumn{1}{c}{\textbf{Generation}} \\
\cmidrule(lr){3-5} 
 &  & \textbf{rFID}$\downarrow$ & \textbf{PSNR$\uparrow$} 
   & \textbf{FID-10K$\downarrow$}  \\
\midrule
$+$ KL $\beta=10^{-6}$ &$10^{-8}$  & 0.39  & 27.12  & 23.12  \\
$+$ KL $\beta=10^{-4}$ &$10^{-7}$ & 0.39  & 27.07  & 23.03  \\
$+$ KL $\beta=10^{-2}$ &$10^{-5}$ & 0.44  & 26.71  & 22.87 \\
$+$ KL $\beta=10^{-1}$ &$10^{-3}$  & 0.50  & 26.23  & 23.16  \\
$+$ KL $\beta=0.5$ &$10^{-2}$ & 0.52  & 26.16  & 22.99 \\
$+$ KL $\beta=1$ &0.07 & 0.61  & 25.45   & 23.18 \\
$+$ KL $\beta=2$ &0.19 & 0.69  & 25.08  & 23.31 \\
$+$ KL $\beta=8$ &0.94 & 2.36  & 22.29  & 27.54 \\
\midrule
$+$ VE Loss  &0.06 &0.46   &26.31   & 18.90 \\
\bottomrule
\end{tabular}}
\label{tab:2}
\end{table}
\subsection{Ablation Study}
In this section, we conduct ablation studies to evaluate the consistent effectiveness of the proposed \textbf{VE} loss in image generation tasks, following the two observations from Section~\ref{sec:4}: (1) modern VAEs are typically trained with a small KL weight, which leads to a latent space that is not robust to diffusion sampling perturbations, and \textbf{VE} can mitigate this issue; (2) tuning the KL regularization coefficient alone cannot fundamentally resolve the problem.

Specifically, to investigate Observation~(1), we consider two widely used baseline tokenizers: a vanilla VAE tokenizer~\cite{rombach2022high}, and a VA-VAE tokenizer~\cite{yao2025reconstruction} that leverages vision foundation models for latent representation learning. For each tokenizer, we compare two variants: one trained with the standard KL objective and the other trained with the proposed \textbf{VE} loss. In these experiments, we train the tokenizer for 80 epochs following the setup of VA-VAE.
To validate Observation~(2), we perform additional experiments by sweeping the KL regularization coefficient. For computational efficiency, we adopt a lightweight VAE trained for 5 epochs and a LightningDiT-B model trained for 80 epochs.

As shown in Table~\ref{tab:1}, VE loss consistently improves the quality of generation under both vanilla and foundation-model-aligned tokenizers, confirming our hypothesis that robust latent representations are crucial for diffusion-based generation.
To further verify the generality of our observations, we conducted extended experiments by training the VA-VAE $+$ VE loss $+$ LightningDiT configuration for 160 epochs to achieve full convergence. The results are compared with those reported in VA-VAE \cite{yao2025reconstruction}, where the VA-VAE was trained for 50 epochs and LightningDiT for 160 epochs.
As shown in Table~\ref{tab:1}, our method achieves better generative performance with only 16 epochs of tokenizer training, surpassing the original VA-VAE trained for 50 epochs. 
These results confirm that the proposed \textbf{VE} loss consistently enhances diffusion performance across different tokenizers, architectures, and training schedules.
More importantly, the robustness and stability of these gains strongly support our central claim: building a latent space that is resilient to diffusion sampling perturbations is fundamental for achieving stable and high-fidelity generation in latent diffusion models.
Moreover, as shown in Table~\ref{tab:2}, the results suggest a limitation of the standard KL objective: its global prior constraint imposes a rigid reconstruction--variance trade-off that is difficult to alleviate by tuning the KL weight alone, whereas our VE loss substantially
mitigates this by adaptively modulating latent variance.
\begin{figure*}[t]
    \centering
    \includegraphics[width=\textwidth]{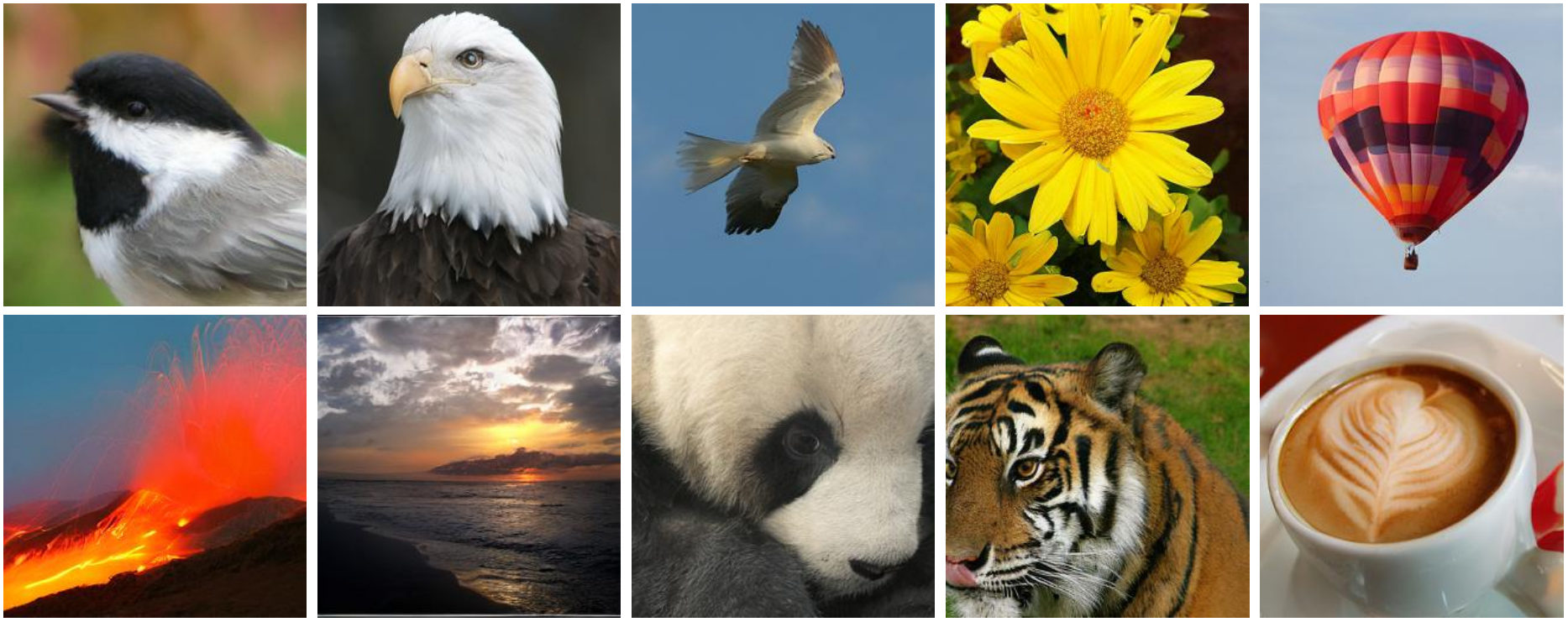}
    \caption{Visualization Results. Examples of class-conditional generation on \textit{ImageNet} 256$\times$256
}
    \label{fig3} 
\end{figure*}
\subsection{Main Results}
We perform a system-level comparison between recent state-of-the-art  approaches, including several autoregressive models~\cite{chang2022maskgit,sun2024autoregressive,VAR,li2024autoregressive,zhang2025mvar} and several diffusion models ~\cite{Zheng2024MaskDiT,peebles2023scalable,ma2024sit,gao2023masked,gao2023mdtv2,yu2024representation,zheng2025diffusion,wu2025representation}.
We use a optimal classifier-free guidance (cfg) scale of 1.70.
Following \cite{yu2024representation,yao2025reconstruction}, we employ cfg interval sampling \cite{kynkaanniemi2024applying}, which has been shown to improve generation quality. We adopt a cfg interval of [0.21, 1]. 
As shown in Table \ref{tab:4}, our method notably reaches a 1.18 FID with only 530 epochs, outperforms the complete state-of-the-art methods, further demonstrating the effectiveness of our approach.
As illustrated in Figure \ref{fig3}, our method produces high-quality images for class-conditional generation on \textit{ImageNet} 256$\times$256. 
Additional qualitative results can be found in Appendix.

\begin{table}[t]
\centering
\caption{System-level comparison on \textit{ImageNet}
256×256 with guidance. All models use classifier-free guidance (CFG) except RAE, which uses auto-guidance. $\downarrow$ and $\uparrow$ indicate whether lower
or higher values are better, respectively.}
\resizebox{\columnwidth}{!}{
\begin{tabular}{l|cc|ccccc}
\toprule
\multirow{2}{*}{\textbf{Method}}  &\multirow{2}{*}{\makecell{\textbf{Training} \\ \textbf{Epoches}}} & \multirow{2}{*}{\textbf{\#params}}  & \multicolumn{5}{c}{\textbf{Generation w/ CFG}} \\
\cmidrule(lr){4-8} 
   & &  & \textbf{gFID}$\downarrow$  & \textbf{sFID}$\downarrow$  & \textbf{IS}$\uparrow$ & \textbf{Pre.}$\uparrow$ & \textbf{Rec.}$\uparrow$ \\
\midrule
LlamaGen~\cite{sun2024autoregressive}  & 300 & 3.1B  & 2.18 & 5.97 & 263.3 & 0.81 & 0.58 \\
VAR~\cite{VAR}   & 350 & 2.0B  & 1.80 & - & 365.4 & 0.83 & 0.57 \\
MVAR~\cite{zhang2025mvar}   & - & 1.0B  & 2.15 & 5.62 & 298.9 & \textbf{0.84} & 0.56 \\
MAR~\cite{li2024autoregressive}   & 800 & 945M & 1.55 & - & 303.7 & 0.81 & 0.62 \\
MaskDiT~\cite{Zheng2024MaskDiT}  & 1600 & 675M  & 2.28 & 5.67 & 276.6 & 0.80 & 0.61 \\ 
DiT~\cite{peebles2023scalable}   & 1400 & 675M  & 2.27 & 4.60 & 278.2 & 0.83 & 0.57 \\
SiT~\cite{ma2024sit}   & 1400 & 675M & 2.06 & 4.50 & 270.3 & 0.82 & 0.59 \\
MDT~\cite{gao2023masked}   & 1300 & 675M  & 1.79 & 4.57 & 283.0 & 0.81 & 0.61 \\
MDTv2~\cite{gao2023mdtv2}   & 1080 & 675M  &1.58 & 4.52 & \textbf{314.7} & 0.79 & 0.65 \\ 
REPA~\cite{yu2024representation}   & 800 & 675M  & 1.42 & 4.70 & 305.7 & 0.80 & 0.65 \\
VA-VAE~\cite{yao2025reconstruction} & 800 & 675M  & 1.35 & \textbf{4.15}  & 295.3 & 0.79 & 0.65 \\
REG~\cite{wu2025representation} & 800 & 675M  & 1.36 & 4.25  & 299.4 & 0.77 & \textbf{0.66} \\
RAE~\cite{zheng2025diffusion} & 800 & 675M  & 1.41 & -  & 309.4 & 0.80 & 0.63 \\
\midrule
\textbf{Ours} &530  &675M  &\textbf{1.18}  &4.29 &289.8   &0.78  &\textbf{0.66}    \\ 
\bottomrule
\end{tabular}}
\label{tab:3}
\end{table}
\begin{table}[t]
\centering
\caption{Reconstruction performance of our finetuned model on \textit{ImageNet}
256×256. $\downarrow$ and $\uparrow$ indicate whether lower
or higher values are better, respectively.}
\resizebox{0.95\columnwidth}{!}{
\begin{tabular}{l|c|cccc}
\toprule
\multirow{2}{*}{\textbf{Method}}  &\multirow{2}{*}{\makecell{\textbf{Training} \\ \textbf{Epoches}}}  & \multicolumn{4}{c}{\textbf{Reconstruction Performance}} \\
\cmidrule(lr){3-6} 
   & & \textbf{rFID}$\downarrow$ & \textbf{PSNR$\uparrow$} & \textbf{LPIPS$\downarrow$} & \textbf{SSIM$\uparrow$} \\
\midrule
VA-VAE~\cite{yao2025reconstruction} & 130 & 0.28  & 27.71 & 0.097 & 0.779 \\
$+$ VE Loss & $+$ 10    &0.26   &28.31  &0.090  &0.792 \\
\bottomrule
\end{tabular}}
\label{tab:4}
\vspace{-2ex}
\end{table}
Table~\ref{tab:3} presents the full reconstruction metrics of our fine-tuned autoencoder. Notably, the baseline VA-VAE training pipeline consists of three stages, where the hyperparameter configuration in Stage~3 is more biased toward reconstruction quality. Since our hyperparameters are mainly aligned with those used in Stage~3, our model achieves slightly better reconstruction than the baseline.
However, this should not be interpreted as \textbf{VE} inherently improving reconstruction. Under a strictly matched training setup (i.e., training from scratch with identical hyperparameters), our method would introduce a slight decrease in reconstruction quality. Nevertheless, the reconstruction performance remains competitive, and the overall generation results are improved, as shown in Tables~\ref{tab:1} and \ref{tab:2}.
Overall, our method enhances robustness to diffusion sampling perturbations while largely preserving reconstruction fidelity. This balance between robustness and reconstruction fidelity is crucial, as it ensures that the improved generative performance is not achieved at the cost of severely degraded reconstructions.

\section{Conclusion}
\label{sec:6}
In this work, we revisit the design of latent spaces for latent diffusion models and reveal that robustness to sampling perturbations is a key factor influencing generative quality, beyond reconstruction accuracy and semantic alignment.
Through both theoretical and empirical analysis, we demonstrate that the conventional KL regularization term in VAE-based tokenizers is not only unnecessary but also detrimental for latent diffusion, as it constrains the representational flexibility of the latent space.
To address this issue, we propose a simple yet effective variance expansion loss that counteracts the variance collapse induced by reconstruction objectives.
By leveraging the natural adversarial interplay between reconstruction compactness and variance expansion, our method adaptively balances latent dispersion and fidelity, resulting in a latent space that is both expressive and robust to stochastic sampling.
Extensive experiments on multiple diffusion baselines validate that the proposed approach consistently improves generation stability and visual quality, while maintaining strong reconstruction performance.
We hope that our findings can provide guidance for designing generative-friendly latent spaces and offer useful insights for future research on improved generative representation learning.

\vspace{5ex}
\noindent\textbf{Acknowledgement}.
This work was supported by National Natural Science Foundation of China (No. 62476051).
\vspace{-2ex}
{
    \small
    \bibliographystyle{ieeenat_fullname}
    \bibliography{main}
}

\clearpage
\appendix     
\setcounter{page}{1}
\maketitlesupplementary

\section{Analyze: Why Gaussian Priors Are Unnecessary in Latent Diffusion.} 
To further understand why an explicit Gaussian prior is not required in latent diffusion models, we begin by revisiting the standard variational autoencoder (VAE) formulation. 
A variational autoencoder models data likelihood through a latent variable \( z \):
\begin{equation}
p_\theta(x) = \int p_\theta(x \mid z)\, p(z)\, \mathrm{d}z,
\end{equation}
where \( p(z) \) is the latent prior, typically chosen as \( \mathcal{N}(0, I) \) for tractability.  
Since the true posterior \( p_\theta(z \mid x) \) is intractable, the encoder learns an approximation \( q_\phi(z \mid x) \), leading to the evidence lower bound (ELBO):
\begin{equation}
\log p_\theta(x)
\ge 
\mathbb{E}_{q_\phi(z \mid x)}[\log p_\theta(x \mid z)]
- \mathrm{KL}\!\left(q_\phi(z \mid x)\,\|\,p(z)\right).
\end{equation}
The Gaussian prior \( p(z)=\mathcal{N}(0,I) \) serves two purposes:
(1) it regularizes the latent space, preventing the encoder from overfitting to individual samples, and 
(2) it allows analytical evaluation of the KL term, enabling stable optimization.
Hence, the Gaussian assumption in VAEs is not merely a modeling choice but a mathematical necessity for tractable variational inference.

In latent diffusion models, the situation is fundamentally different.  
The latent variable \( z_0 \) (obtained from a tokenizer or encoder) is further diffused through a forward noising process:
\begin{equation}
q(z_t \mid z_{t-1}) = \mathcal{N}(\sqrt{\alpha_t} z_{t-1}, (1-\alpha_t)I),
\end{equation}
and the model learns the reverse transitions \( p_\theta(z_{t-1} \mid z_t) \).  
This process defines an \emph{implicit prior} over \( z_0 \):
\begin{equation}
p_\theta(z_0) = \int p(z_T) \prod_{t=1}^T p_\theta(z_{t-1}\mid z_t)\,\mathrm{d}z_{1:T},
\end{equation}
where \( p(z_T) = \mathcal{N}(0,I) \) is only the noise prior at the terminal step.  
The marginal distribution \( p_\theta(z_0) \) over the encoder’s latent space is therefore \emph{learned} by the diffusion model itself rather than fixed a priori.  
Consequently, constraining \( q_\phi(z_0 \mid x) \) to follow a Gaussian distribution is both unnecessary and potentially harmful, as it restricts the expressiveness and robustness of the latent space learned through diffusion.

In summary, the key distinction between VAEs and latent diffusion models lies in where and how the latent prior is defined. 
While VAEs rely on an explicit, analytically specified Gaussian prior to regularize the latent distribution, latent diffusion models instead learn an implicit prior through the denoising trajectory and its associated score (or velocity) field.  
As a consequence, enforcing a predefined Gaussian constraint on the encoder’s output is not only unnecessary, but can also distort the intrinsic geometry of the latent manifold, reduce its expressiveness, and ultimately hinder the diffusion model’s ability to learn a faithful, data-driven prior in the latent space.

\section{More Training Details}
\label{sec:b}
For diffusion models trained in latent spaces aligned with DINOv2~\cite{oquab2023dinov2} representations~\cite{yao2025reconstruction,zheng2025diffusion}, it is well known that, despite their strong performance, they tend to suffer from training instabilities under long optimization schedules. In practice, this often manifests as sudden loss spikes in the later stages of training, after which the optimization rarely recovers—a phenomenon that has been widely reported in the community. To mitigate this issue, RAE~\cite{zheng2025diffusion} employs a learning rate schedule that linearly decays from $2.0 \times 10^{-4}$ to $2.0 \times 10^{-5}$ with a constant warmup of 40 epochs. In our experiments, we observed that using the Muon optimizer~\cite{jordan2024muon} substantially alleviates this issue. Therefore, for all long-horizon training runs in this work, we adopt Muon as our default optimizer.

\section{Details of the 2D toy example}
\label{sec:c}
We largely follow the dataset construction protocol of Karras et al.~\cite{karras2024guiding}, with one important modification: since our experiments do not make use of any class-dependent effects, we restrict the data distribution to a single class. 

More specifically, we represent the fractal-like structure of the data by a Gaussian mixture
\mbox{$\mathcal{M}_{\mathbf{c}} = \big( \{\phi_i\}, \{\boldsymbol{\mu}_i\}, \{\boldsymbol{\Sigma}_i\} \big)$},
where $\phi_i$, $\boldsymbol{\mu}_i$, and $\boldsymbol{\Sigma}_i \in \mathbb{R}^{2 \times 2}$ denote the mixture weight, the mean, and the covariance matrix of component $i$, respectively. This parameterization admits closed-form expressions for both the density and its score, which enables exact computation and visualization without any further approximations. For a fixed class $\mathbf{c}$, the data density can be written as
\begin{equation}
p_{\text{data}}(\mathbf{x}\mid \mathbf{c})
= \sum_{i \in \mathcal{M}_{\mathbf{c}}} \phi_i \,
\mathcal{N}(\mathbf{x}; \boldsymbol{\mu}_i, \boldsymbol{\Sigma}_i),
\end{equation}
where the two-dimensional Gaussian density is given by
\begin{equation}
\mathcal{N}(\mathbf{x}; \boldsymbol{\mu}, \boldsymbol{\Sigma})
= \frac{1}{2\pi \sqrt{\det(\boldsymbol{\Sigma})}}
   \exp\Big(
   -\tfrac{1}{2}(\mathbf{x}-\boldsymbol{\mu})^{\top}
   \boldsymbol{\Sigma}^{-1} (\mathbf{x}-\boldsymbol{\mu})
   \Big).
\end{equation}

Adding isotropic Gaussian noise of standard deviation $\sigma$ to $p_\text{data}(\mathbf{x} \mid \mathbf{c})$ corresponds to convolving it with a Gaussian kernel, which yields a family of smoothed densities $p(\mathbf{x} \mid \mathbf{c}; \sigma)$ parameterized by the noise level:
\begin{equation}\label{eq:GT}
p(\mathbf{x}\mid \mathbf{c}; \sigma)
= \sum_{i \in \mathcal{M}_{\mathbf{c}}} \phi_i \,
\mathcal{N}(\mathbf{x}; \boldsymbol{\mu}_i, \boldsymbol{\Sigma}_{i,\sigma}^*),
\quad\text{with}\quad
\boldsymbol{\Sigma}_{i,\sigma}^* = \boldsymbol{\Sigma}_i + \sigma^2 \mathbf{I}.
\end{equation}
The corresponding score function admits the closed-form expression
\begin{equation}
\nabla_{\mathbf{x}} \log p(\mathbf{x}\mid \mathbf{c}; \sigma)
=
\frac{
\sum_{i \in \mathcal{M}_{\mathbf{c}}}
\phi_i \,
\mathcal{N}(\mathbf{x}; \boldsymbol{\mu}_i, \boldsymbol{\Sigma}_{i,\sigma}^*)
\, (\boldsymbol{\Sigma}_{i,\sigma}^*)^{-1} (\boldsymbol{\mu}_i - \mathbf{x})
}{
\sum_{i \in \mathcal{M}_{\mathbf{c}}}
\phi_i \,
\mathcal{N}(\mathbf{x}; \boldsymbol{\mu}_i, \boldsymbol{\Sigma}_{i,\sigma}^*)
}
.
\end{equation}

To obtain a thin, tree-shaped structure, we design $\mathcal{M}_{\mathbf{c}}$ by starting from a single main “branch” and recursively splitting it into smaller sub-branches. Each branch segment is represented by 8 anisotropic Gaussian components. The subdivision is repeated 6 times; after each split, we downscale the corresponding mixture weights $\phi_i$ and introduce small random perturbations to the lengths and orientations of the two child branches. This procedure produces
\mbox{$127 \times 8 = 1016$} components for the class considered in our experiments. Following the normalization guidelines of Karras et al.~\cite{karras2024guiding}, we choose the coordinate system so that the mean of $p_\text{data}$ (marginalized over $\mathbf{c}$) is zero and the standard deviation along each axis is $\sigma_\text{data} = 0.5$.

\paragraph{Models and training details.}
We implement both the tokenizer and the denoiser (vector-field) networks as 8-layer ReLU MLPs (hidden dim 512).
To make the latent space directly visualizable, we fix its dimensionality to two. Concretely, the tokenizer encoder maps a two-dimensional input point $\boldsymbol{x}\in\mathbb{R}^2$ to a two-dimensional latent code $\boldsymbol{z}\in\mathbb{R}^2$, and the decoder maps it back to the data space:
\begin{align}
    \boldsymbol{z} &= f_{\text{enc}}(\boldsymbol{x}; \boldsymbol{\theta}_{\text{enc}}) \in \mathbb{R}^2, \\
    \hat{\boldsymbol{x}} &= f_{\text{dec}}(\boldsymbol{z}; \boldsymbol{\theta}_{\text{dec}}) \in \mathbb{R}^2.
\end{align}
The tokenizer is trained using a standard reconstruction objective with KL regularization or VE loss: 
\begin{equation}
    \mathcal{L}_{\text{rec}}
    = \mathbb{E}_{\boldsymbol{x} \sim p_{\text{data}}}
      \big[\,\|\boldsymbol{x} - f_{\text{dec}}(f_{\text{enc}}(\boldsymbol{x}))\| \big].
\end{equation}
In Figure \ref{fig1}, the baseline uses a KL coefficient of $10^{-3}$, while VE loss uses a coefficient of $10^{-2}$.

For the diffusion model, we adopt the flow-matching formulation, which provides a particularly simple and effective way to learn continuous-time generative dynamics. We define a bridging trajectory between a simple base distribution $p_0$ and the data distribution $p_{\text{data}}$ as
\begin{equation}
    \boldsymbol{Z}_t = (1-t)\,\boldsymbol{Z}_0 + t\,\boldsymbol{X},
    \quad t \in [0,1],
\end{equation}
where $\boldsymbol{Z}_0 \sim p_0$ (we use a standard Gaussian $\mathcal{N}(\boldsymbol{0}, \boldsymbol{I})$) and $\boldsymbol{X} \sim p_{\text{data}}$. The corresponding ground-truth velocity is simply
\begin{equation}
    \dot{\boldsymbol{Z}}_t = \frac{\mathrm{d}\boldsymbol{Z}_t}{\mathrm{d}t}
    = \boldsymbol{X} - \boldsymbol{Z}_0.
\end{equation}
Flow-matching models directly learn the time-dependent vector field $v_{\boldsymbol{\theta}}$ that approximates this velocity along the trajectory, using the loss
\begin{equation}
\mathcal{L}_{\text{flow}} 
= \mathbb{E}_{t \sim \mathcal{U}(0,1),\,\boldsymbol{Z}_t}
\big[ \| v_{\boldsymbol{\theta}}(\boldsymbol{Z}_t, t) - \dot{\boldsymbol{Z}}_t \|_2^2 \big],
\end{equation}
where $v_{\boldsymbol{\theta}}$ denotes the predicted instantaneous velocity and $\dot{\boldsymbol{Z}}_t$ is the ground-truth time derivative of the path defined above.

Both the tokenizer and the flow-matching model are trained with Adam for 200k iterations using batch size 4096, learning rate $10^{-3}$ with a schedule following AutoGuidance.
At sampling time, we uses an Euler solver with 20 steps:
\begin{equation}
    \frac{\mathrm{d}\boldsymbol{Z}_t}{\mathrm{d}t} = v_{\boldsymbol{\theta}}(\boldsymbol{Z}_t, t),
\end{equation}
starting from $\boldsymbol{Z}_0 \sim p_0$ and evolving from $t=0$ to $t=1$ using a standard explicit Euler sampler with $N=20$ uniform time steps. Denoting $t_k = k/N$ and $\Delta t = 1/N$, the sampler update reads
\begin{align}
    \boldsymbol{Z}_{t_{k+1}}
    &= \boldsymbol{Z}_{t_k}
       + \Delta t \, v_{\boldsymbol{\theta}}(\boldsymbol{Z}_{t_k}, t_k),
       \quad k = 0,\dots, N-1,
\end{align}
and the final sample in data space is obtained by decoding the terminal latent state,
\begin{equation}
    \hat{\boldsymbol{x}} = f_{\text{dec}}(\boldsymbol{Z}_{t_1}).
\end{equation}
This setup keeps the model and training procedure minimal while allowing us to directly inspect both the learned latent representation and the generative trajectories in two dimensions.

\section{Visual Comparison}
\label{sec:G}
\begin{figure*}[h]
    \vspace{5ex}
    \centering
    \includegraphics[width=\textwidth]{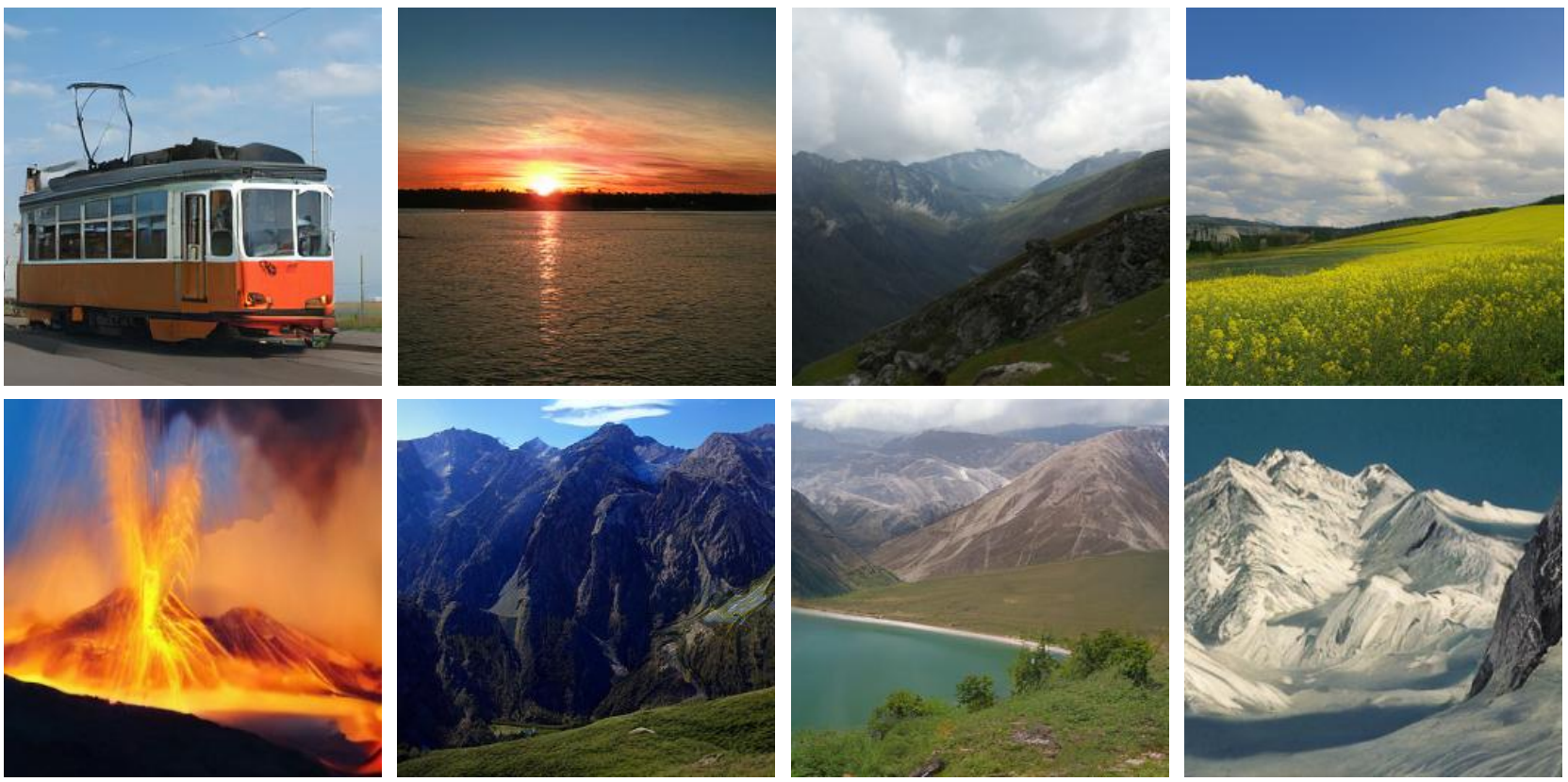}
    \caption{Visualization Results. Examples of class-conditional generation on \textit{ImageNet} 256$\times$256
}
    \label{sub1} 
\end{figure*}

\begin{figure*}[t]
    \centering
    \includegraphics[width=\textwidth]{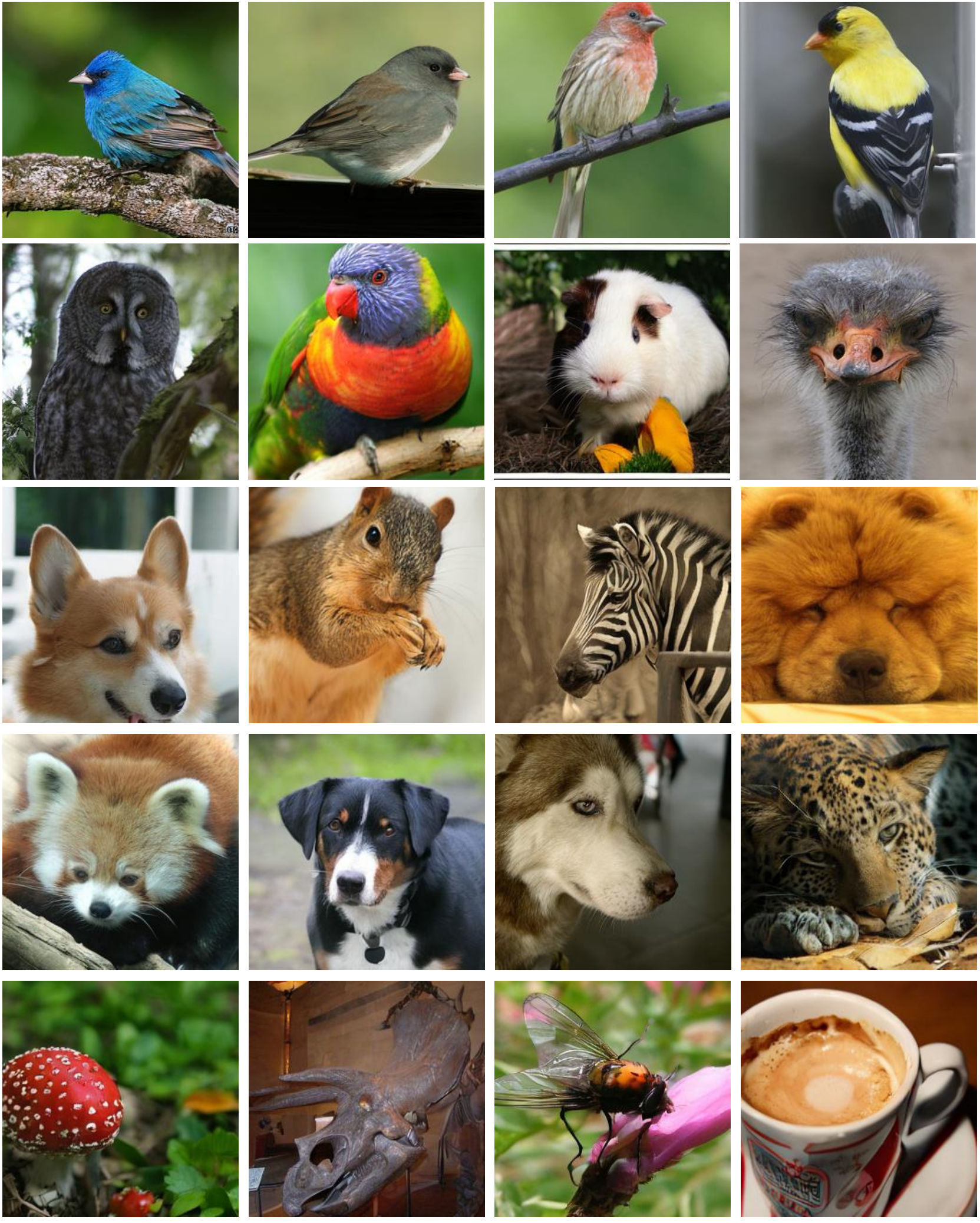}
    \caption{Visualization Results. Examples of class-conditional generation on \textit{ImageNet} 256$\times$256
}
    \label{sub2} 
\end{figure*}

We provide additional visual examples of our method on \textit{ImageNet} $256 \times 256$. 
Consistent with Figure~\ref{fig3} and Table~\ref{tab:4}, all samples are generated with a classifier-free guidance (CFG) scale of $1.45$ and a CFG interval of $[0.13, 1]$. 
Representative visual results are shown in Figures~\ref{sub1} and~\ref{sub2}.


\end{document}